\documentclass[a4paper]{article}
\usepackage{cite}
\usepackage{graphicx, subfigure}
\usepackage{float}
\usepackage[cmex10]{amsmath}
\usepackage{amssymb,amsthm,bm}
\usepackage{enumerate}
\usepackage{epstopdf}
\usepackage{url}
\usepackage{geometry}
\usepackage{multirow}
\usepackage{algorithm, algorithmic}
\usepackage{underscore} 

\title{Time Series Compression Based on Adaptive Piecewise Recurrent Autoencoder}
\author{Daniel Hsu\\
\\
School of Electrical and Computer Engineering\\
Georgia Institute of Technology\\
Atlanta, Georgia 30332--0250\\}

\begin{document}

\maketitle

\begin{abstract}
Time series account for a large proportion of the data stored in financial, medical and scientific databases. The efficient storage of time series is important in practical applications. In this paper, we propose a novel compression scheme for time series. The encoder and decoder are both composed by recurrent neural networks (RNN) such as long short-term memory (LSTM). There is an autoencoder between encoder and decoder, which encodes the hidden state and input together and decodes them at the decoder side. Moreover, we pre-process the original time series by partitioning it into segments with various lengths which have similar total variation. The experimental study shows that the proposed algorithm can achieve competitive compression ratio on real-world time series.
\end{abstract}

\section{Introduction}
Time-series data is one of the most important types of data, and it is increasingly collected in many different domains \cite{rlg}. The high-resolution data would be extremely beneficial for many analytical applications such as data visualization \cite{dv}, short-term forecasts of energy generation/load \cite{forecast} or energy prices \cite{price}. The explanation for not working with high-resolution data is the enormous amount of storage space required. For example, in electrical grid, utility companies serving millions of customers, storing smart-meter measurements of high resolution would sum up to petabytes of data \cite{8}. Even if storage prices are expected to decrease, the benefits of analytical applications will not always justify huge investments in storage. Similar concerns hold for the transmission of such data, which might cause congestion.

To deal with such huge amounts of fine-grained time series data, one possibility is to employ compression techniques. In contrast to lossless compression techniques \cite{35} such as \cite{13}\cite{45}, lossy compression promises to achieve much higher compression ratios \cite{salomon}. When doing lossy compression of data, one has to ensure that the original data can be retrieved or reconstructed with sufficient quality for the respective applications. This is, tasks like precise short-term load forecasting can only be done if very-close-to-original measurement values can be reconstructed from the compressed data. This requires guarantees regarding the maximum deviation of the values. Although time-series data and lossless compression have been investigated for quite some time, there is little research on lossy compression and retrieval of such data with guarantees. Studies such as \cite{9}\cite{22}\cite{31} do not compress all types of data well or have disadvantages in terms of runtime. Moreover, none of these articles propose methods estimating the performance of their compression techniques.

RNNs are particularly suitable for modeling dynamical systems as they operate on input information as well as a trace of acquired previous information (due to recurrent connections) allowing for direct processing of temporal dependencies. RNNs can be employed for a wide range of tasks as they inherit their flexibility from plain neural networks. Among all RNN architectures, the most successful one to characterize long-term memory is the long short-term memory network (LSTM) \cite{lstm}, which learns both short-term and long-term memory by enforcing constant error flow through the designed cell state. 

In this work, we propose a novel scheme for lossy compression of time series. We model the encoder and decoder by two LSTM cells working parallely. Between encoder and decoder, we use the autoencoder to compress the hidden state and input observations together. Since the LSTM has strong capability of learning long-term dependencies, the encoder and decoder can both capture the long-term dependencies in time series, which can reduce the amount of information needed in the transmission. Because the input size of LSTM and autoencoder is fixed, we need to an extra interpolation step to adapt to the changes of local statistics in time series, which can significantly reduce the compression efficiency. And the proposed algorithm can be easily extended to multi-dimensional time series. The experimental study shows the compression capability of the proposed algorithm.

\section{Preliminary}
\subsection{Autoencoder}
The Autoencoder (AE) was first introduced as a dimension-reduction model \cite{ae}. An autoencoder takes an input vector $\bm{x}$ and transforms it into a latent representation $\bm{z}$. The transformation, typically referred  as encoder, follows the equation as below:
\begin{equation}
\bm{z}=\sigma(\bm{W}\bm{x}+\bm{b}) \nonumber
\end{equation}
where $\bm{W}$ and $\bm{b}$ correspond to the weighs and bias in the encoder, and $\sigma$ is the sigmoid function.

The resulting latent representation $\bm{z}$ is then mapped back into the reconstructed feature space $\bm{y}$ in the decoder as follows:
\begin{equation}
\bm{y}=\sigma(\bm{W}'\bm{z}+\bm{b}') \nonumber
\end{equation}
where $\bm{W}'$ and $\bm{b}'$ correspond to the weights and bias in the decoder. The autoencoder is trained by minimizing the reconstruction error $\|\bm{y}-\bm{x}\|$. 

\subsection{Recurrent Neural Network}
RNNs are discrete-time state–space models trainable by specialized weight adaptation algorithms. The input to RNN is a variable-length sequence $\bm{x}=(\bm{x}_1,\ldots,\bm{x}_T)$ which can be recursively processed. And when processing each symbol, RNN maintains its internal hidden state $\bm{h}$. The operation of RNN at each timestep $t$ can be formulated as
\begin{equation}
\bm{h}_t=f_{\theta}(\bm{x}_t, \bm{h}_{t-1}) \nonumber
\end{equation}
where $f$ is the deterministic state transition function and $\theta$ is the parameter of $f$. The output of RNN is computed using the following equation:
\begin{equation}
\bm{y}_t=g_{\varphi}(\bm{h}_t)
\end{equation}
where function $g$ can be modeled as a neural network with weights $\varphi$. In implementation, the function $f$ can be realized by long short-term memory \cite{lstm}. 

Although traditional RNN exhibits a superior capability of modeling nonlinear time series problems in an effective fashion, there are still several issues to be addressed \cite{ger}:
\begin{itemize}
\item Traditional RNNs cannot train the time series with very long time lags, which is commonly seen in real-world datasets.
\item Traditional RNNs rely on the predetermined time lags to learn the temporal sequence processing, but it is difficult to find the optimal window size in an automatic way.
\end{itemize}

\subsection{Long Short-Term Memory}
To overcome the aforementioned disadvantages of traditional RNNs, Long Short-Term Memory (LSTM) neural network is adopted in this study to model time series. LSTM was initially introduced in \cite{lstm} with the objective of modeling long-term dependencies and determining the optimal time lag for time series problems. A LSTM is composed of one input layer, one recurrent hidden layer, and one output layer. The basic unit in the hidden layer is memory block, containing memory cells with self-connections memorizing the temporal state, and a pair of adaptive, multiplicative gating units to control information flow in the block. It also has input gate and output gate controlling the input and output activations into the block. 

The memory cell is primarily a recurrently self-connected linear unit, called Constant Error Carousel (CEC), and the cell state is represented by the activation of the CEC. Because of CEC, the multiplicative gates can learn when to open and close. Then by keeping the network error constant, the vanishing gradient problem can be solved in LSTM. Moreover, a forget gate was added to the memory cell, which can prevent the gradient from exploding when learning long time series. The operation and structure of LSTM can be described as below:
\begin{eqnarray}
\bm{i}_t&=&\sigma(\bm{W}_i\bm{x}_t+\bm{U}_i\bm{m}_{t-1}+\bm{b}_i) \nonumber \\
\bm{o}_t&=&\sigma(\bm{W}_o\bm{x}_t+\bm{U}_o\bm{m}_{t-1}+\bm{b}_o) \nonumber \\
\bm{f}_t&=&\sigma(\bm{W}_f\bm{x}_t+\bm{U}_f\bm{m}_{t-1}+\bm{b}_f) \nonumber \\
\bm{c}_t&=&\bm{f}_t\circ\bm{c}_{t-1}+\bm{i}_t\circ\text{tanh}(\bm{W}_c\bm{x}_t+\bm{U}_c\bm{m}_{t-1}+\bm{b}_c) \nonumber \\
\bm{m}_t&=&\bm{o}_t\circ\text{tanh}(\bm{c}_t) \label{state}
\end{eqnarray}
where $\bm{i}_t, \bm{f}_t$ and $\bm{o}_t$ are denoted as input gate, forget gate and output gate at time $t$ respectively, $\bm{m}_t$ and $\bm{c}_t$ represent the hidden state and cell state of the memory cell at time $t$, and $\circ$ is the elementwise multiplication.

\section{Model: Adaptive Pairwise Recurrent Autoencoder}
The model presented in this work is a combination of autoencoder and LSTM, where the input window size can also adjust to local statistics of time series. Denote the input observations, compressed signal, and reconstruction at time $t$ as $\bm{x}_t, \bm{h}_t$ and $\hat{\bm{x}}_t$ respectively. The input window size, i.e. the dimension of vector $\bm{x}_t$, is denoted as $l_t$.

\subsection{Recurrent Autoencoder}
On the encoder side, at each time step $t$, the encoder reads the input $\bm{x}_t$, and extracts its feature $\bm{z}_t:=\varphi_x(\bm{x}_t)$. Inside the LSTM cell, the state transition can be described as below:
\begin{equation}
\bm{c}_{t+1}, \bm{m}_{t+1} = f^{\text{enc}}_{\theta}(\bm{z}_t, \bm{c}_t, \bm{m}_t) \label{lstm}
\end{equation}
where the function $f$ is the operation described by \eqref{state} parametrized by $\theta$. The compressed signal is calculated by the output function $g^{\text{enc}}$ as below:
\begin{equation}
\bm{h}_t=g^{\text{enc}}(\bm{z}_t, \bm{m}_{t-1}) \label{enc}
\end{equation}
where output function is implemented by a two-layer neural network. Since LSTM cell at the decoder can also capture the long-term dependencies, we don't need encode cell state $\bm{c}_t$ into compressed signal. The incorporation of the previous hidden state $\bm{m}_{t-1}$ is not redundant, because that hidden state contains information to predict the current input $\bm{x}_t$ which can improve the compression efficiency. Actually, the popular predictive coding compression is to compress the prediction error, i.e., $\bm{x}_t-\hat{\bm{x}}_t(\bm{h}_{t-1})$. In high dimension case, the prediction residual is not small, where the prediction based compression may not work well. However, compressing the input signal together with the previous hidden state can augment the compression, and predictive coding is a special case of our method.

On the decoder side, denote the hidden state and cell state of LSTM as $\bm{m}'_t$ and $\bm{c}'_t$. When receiving the compressed signal, the decoder first decodes the input feature and memory state:
\begin{equation}
\hat{\bm{z}}_t, =g^{\text{dec}}(\bm{h}_t) \nonumber
\end{equation}
where $g^{\text{dec}}$ is also implemented by two-layer neural network. The LSTM state transition at the decoder side is described as:
\begin{equation}
\bm{c}'_{t+1}, \bm{m}'_{t+1}=f^{\text{dec}}_{\xi}(\hat{\bm{z}}_t, \bm{m}'_t) \label{lstm-dec}
\end{equation}
The reconstructed signal is obtained by the output function at the decoder:
\begin{equation}
\hat{\bm{x}}_t=o(\varphi_z(\hat{\bm{z}}_t), \bm{c}'_t, \bm{m}'_t) \nonumber
\end{equation}
where function $o$ is implemented by two-layer neural network, $\bm{m}_t$ here is the hidden state of the decoder LSTM cell, and $\varphi_z$ is the inverse function of $\varphi_x$ at the encoder side. Since both the encoder and decoder can learn the long-term dependencies at the same time, the dimension of compressed signal $\bm{h}_t$ can be significantly reduced. The whole structure is shown in Figure 1.
\begin{figure}[H]
\centering
\includegraphics[width=2.7in]{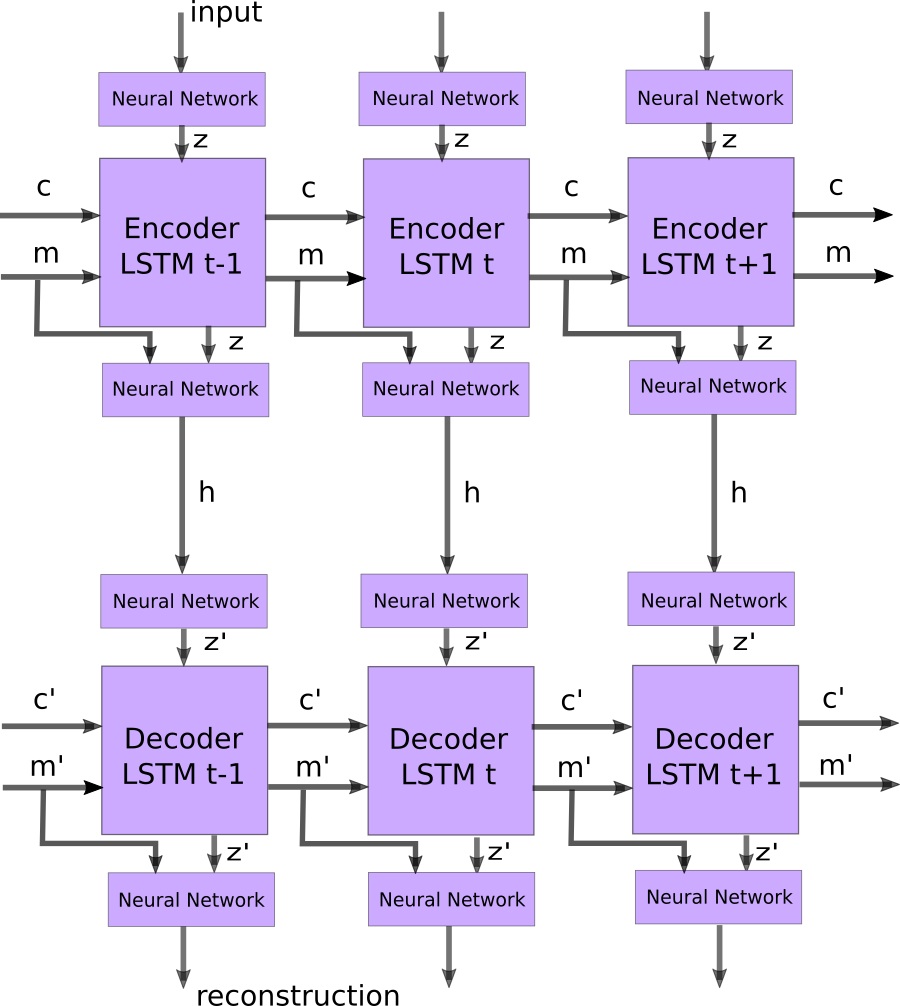}
\caption{The Unrolled Structure of Recurrent Autoencoder (RAE).}
\end{figure}

\subsection{Adaptive Pairwise Compression Algorithm}
Our piecewise compression algorithm employs a greedy strategy to compress intervals of time series. This is necessary as the size of the state space for an optimal solution is extremely large. 
 
For most kinds of time series, such as seismic signals, the local statistics are changing dramatically across time. It can be smooth and flat in some time, but it can also be fluctuating and has great variance in some time. According to \cite{learn}, the learnability of fully connected neural network is constraint by the variance of the function to learn, which is the $\mathit{l}$-1 norm of the coefficients of the function represented by 1-Lipschitz basis. For simplicity, we preprocess the training data based on local total variation, i.e.,
\begin{equation}
\Delta(\mathcal{L})=\sum_{t\in\mathcal{L}}|\bm{x}_t-\bm{x}_{t-1}| \nonumber
\end{equation}
where $\mathcal{L}$ is the local time window. During training, for each trace, we greedily partition time series into segments, where the total variation of each segments is close to a pre-defined value $\tau$ which is specific to the property of dataset. When the length of the segment $|\mathcal{L}|$ is greater than the input dimension of RAE $d_{\text{in}}$, this segment is down sampled to $d_{\text{in}}$, since this segment is smooth and has redundant information. When the length of the segment $|\mathcal{L}|$ is smaller than $d_{\text{in}}$, we interpolate this segment to the length of $d_{\text{in}}$, because the interpolated of this segment is easier to learn. This pre-processing facilitate the training without losing important information. The following figures show two seismic traces before and after pre-processing.

\begin{figure}[H]
\begin{minipage}[t]{0.5\linewidth}
\centering
\includegraphics[width=2.3in]{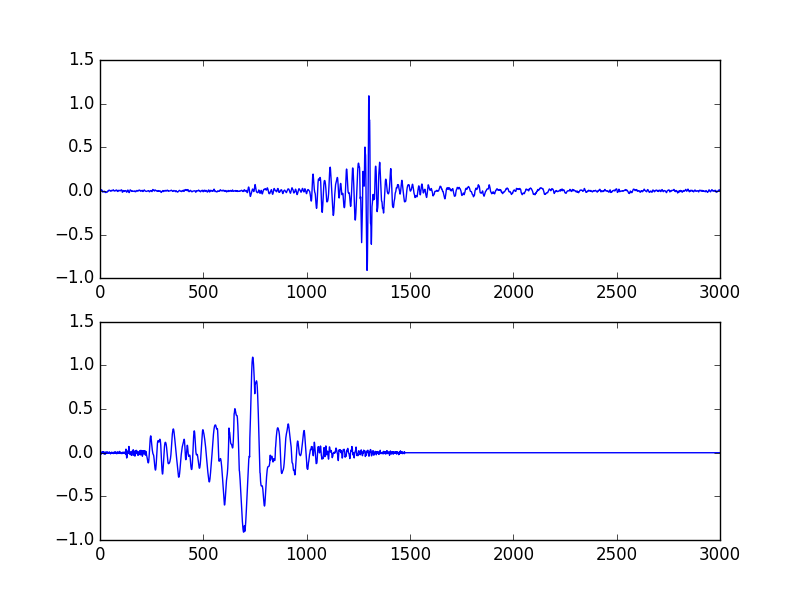}
\label{fig:side:a}
\end{minipage}%
\begin{minipage}[t]{0.5\linewidth}
\centering
\includegraphics[width=2.3in]{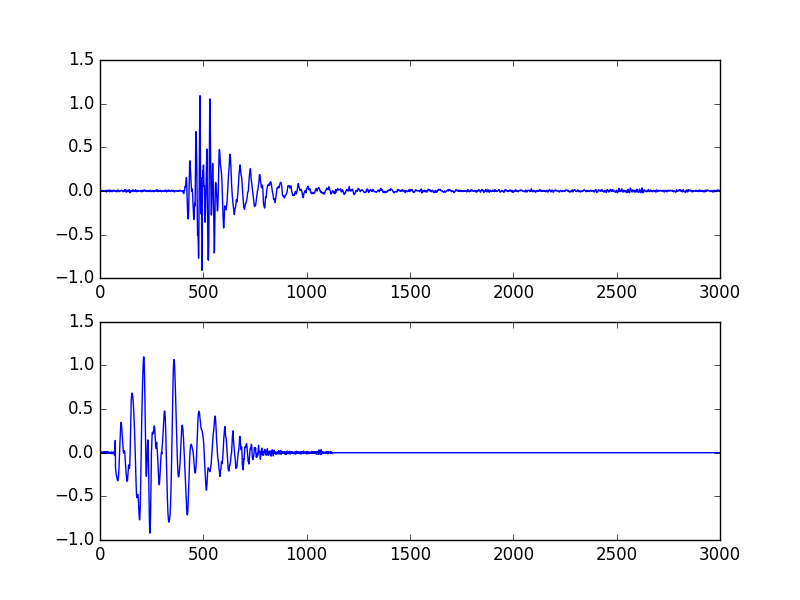}
\label{fig:side:b}
\end{minipage}
\caption{The effect of pre-processing on seismic traces: threshold $\tau=0.4$.}
\end{figure}

After trained on pre-processed dataset, the recurrent autoencoder (RAE) can be a good compressor for each segment of time series. The main algorithm (Algorithm 1) of our compression algorithm employs RAE incrementally, and the compression result depends on a user-defined maximum tolerable deviation. This deviation is realized as a threshold on the uniform norm between the original and the reconstructed time series. This norm is defined as the maximum absolute distance between any pair of points of the real $\bm{x}_t$ and the reconstructed $\hat{\bm{x}}_t$ time series:
\begin{equation}
L_{\infty}=\max_{t=1,\ldots,n}|\bm{x}_t-\hat{\bm{x}}_t| \label{linf}
\end{equation}
In order to improve the compression efficiency, the size of input observations is changed according to the local statistics of time series. So, the number of input observations may be different from the input dimension of RAE. When the nearby data points are relatively smooth, which is easier to compress, the input window size can be larger and more data are compressed at once. In this case, the input observations are down-sampled to meet the input dimension of RAE. When the nearby data points are changing with higher diversity, which is relatively difficult to compress, the input window size is decrease. Here, in order to meet the input dimension of RAE, the input observations are interpolated to longer vector. 

Therefore, the time series are encoded with various resolution. Some parts are compressed in a coarse way while some parts are compressed with more refine codes, which can significantly improve the compression efficiency. In implementation, at each time, we use binary search to find the best input window size.
\begin{algorithm}[H]
\caption{}
\label{alg1}
\begin{algorithmic}[1]
\STATE Let $S$ be the time series for compression.
\STATE Let \emph{data_len}, \emph{rae_len}, \emph{input_len} and \emph{len_stride} denote the length of time series, the input dimension of RAE, input window size, and increment stride during the length search respectively.
\STATE Let \emph{st} denote the starting point of current interval of time series.
\STATE Let \emph{interpolate}() denote the operation which interpolates or down-samples the input vector to have the length of \emph{rae_len}.
\STATE \emph{st} = 0
\WHILE{$\text{\emph{st}}+\text{\emph{rae_len}}<\text{\emph{data_len}}$}
\STATE $\text{\emph{input_len}}=\text{\emph{data_len}}-\text{\emph{st}}$
\STATE $\text{\emph{len_stride}}=\text{\emph{input_len}}/2$
\WHILE{\emph{st}+\emph{input_len} is not going to exceed \emph{data_len} and $\text{\emph{len_stride}}>1$}
\STATE $\bm{x}_t=S[\text{\emph{st}}:\text{\emph{st}}+\text{\emph{input_len}}]$
\STATE $\tilde{\bm{x}}_t=\text{\emph{interpolate}}(\bm{x}_t)$
\STATE Feed $\tilde{\bm{x}}_t$ into the trained RAE model, and get reconstructed $\hat{\bm{x}}_t$.
\STATE Calculate the max deviation $L_{\infty}$ as \eqref{linf}. 
\IF{$L_{\infty}$ is smaller than user-defined max tolerable deviation}
\STATE Record \emph{input_len}.
\STATE $\text{\emph{input_len}}=\text{\emph{input_len}}+\text{\emph{len_stride}}$
\ELSE
\STATE $\text{\emph{input_len}}=\text{\emph{input_len}}-\text{\emph{len_stride}}$
\ENDIF
\STATE $\text{\emph{len_stride}}=\text{\emph{len_stride}}/2$
\ENDWHILE
\STATE Recover the last recorded \emph{input_len}.
\STATE $\bm{x}_t=S[\text{\emph{st}}:\text{\emph{st}}+\text{\emph{input_len}}]$
\STATE Feed the \emph{interpolate}($\bm{x}_t$) into RAE, and get reconstructed $\hat{\bm{x}}_t$.
\STATE Save $\hat{\bm{x}}_t$ as an interval of reconstructed time series $S$.
\STATE $\text{\emph{st}}=\text{\emph{st}}+\text{\emph{input_len}}$
\ENDWHILE
\end{algorithmic}
\end{algorithm}

\section{Experiment}
In this section, we conduct experimental study to show the compression capability of the proposed algorithm. 
\subsection{1D Time Series}
In this experiment, the time series are from seismic signal \cite{seismic}. We first train the RAE model on a qualified seismic dataset with malicious traces evicted. Then we apply the adaptive pairwise compression algorithm on the trained RAE model. All traces in training and test sets are normalized to [-1, 1]. For each signal trace, the tolerable deviations 0.1 and 0.15 are tested. The dashed red lines represent the boundaries between adaptive input windows. The green line shows the reconstructed signal, and the blue shows the original signal.

\begin{figure}[H]
\centering
\includegraphics[width=6in]{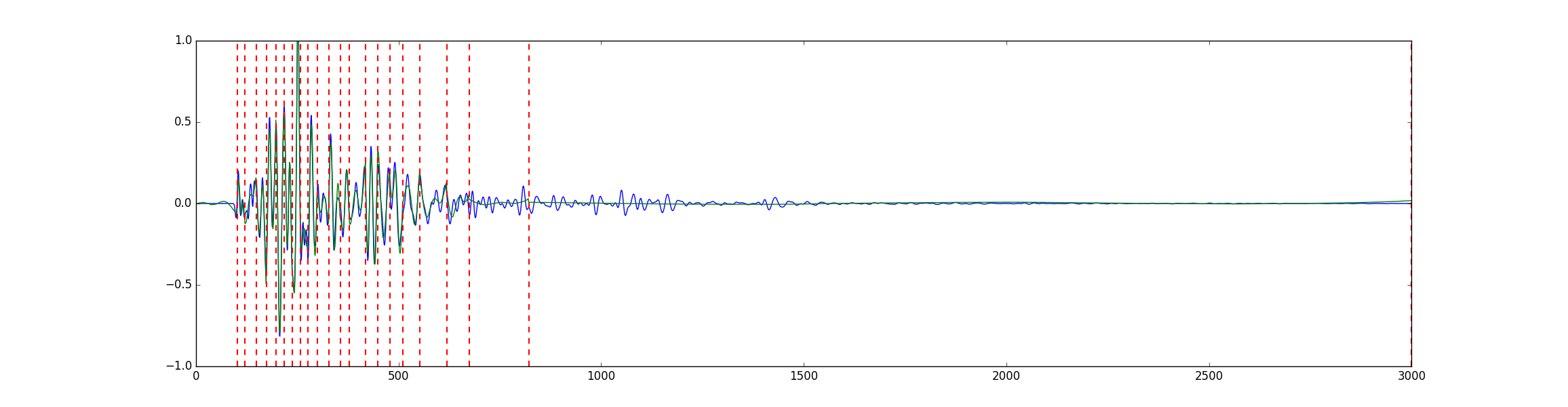}
\caption{Max tolerable deviation:0.1 Compression ratio:0.06 RMSE: 0.03 }
\end{figure}
\begin{figure}[H]
\centering
\includegraphics[width=6in]{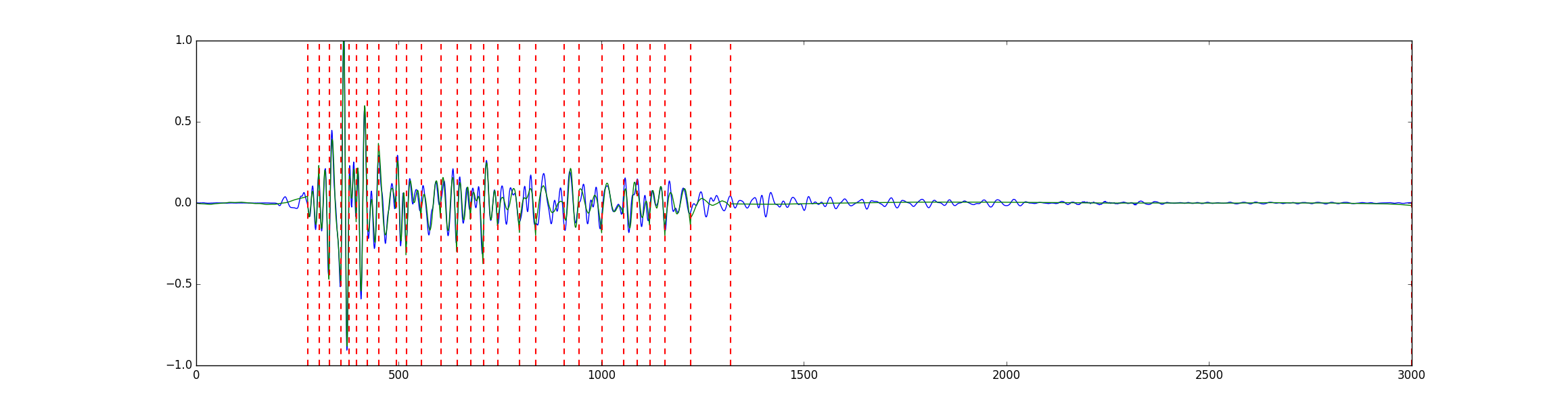}
\caption{Max tolerable deviation:0.1 Compression ratio:0.08 RMSE: 0.02}
\end{figure}
\begin{figure}[H]
\centering
\includegraphics[width=6in]{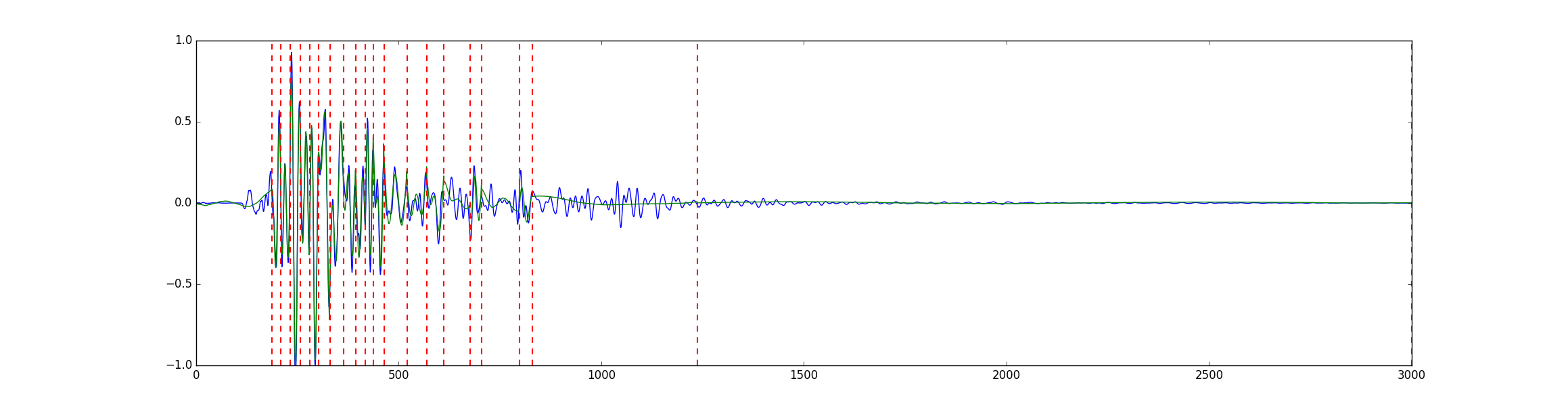}
\caption{Max tolerable deviation:0.15 Compression ratio:0.05 RMSE: 0.05 }
\end{figure}
\begin{figure}[H]
\centering
\includegraphics[width=6in]{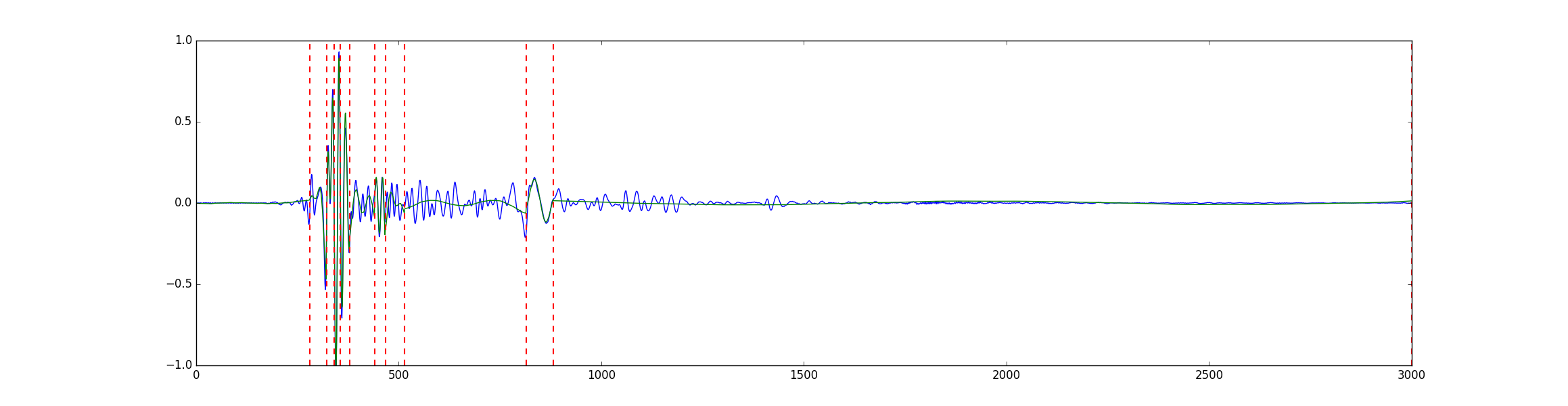}
\caption{Max tolerable deviation:0.15 Compression ratio:0.03 RMSE: 0.07}
\end{figure}

\subsection{Multi-dimensional Time Series}
In the second experiment, we evaluate the proposed method on human activity time series from smartphone sensors \cite{fusion}. The signal has three dimensions, X, Y and Z-axis. The training was performed on the first 1000 data points, and the compression was evaluated on next 1000 data points. The results are shown below.
\begin{figure}[H]
\centering
\includegraphics[width=4in]{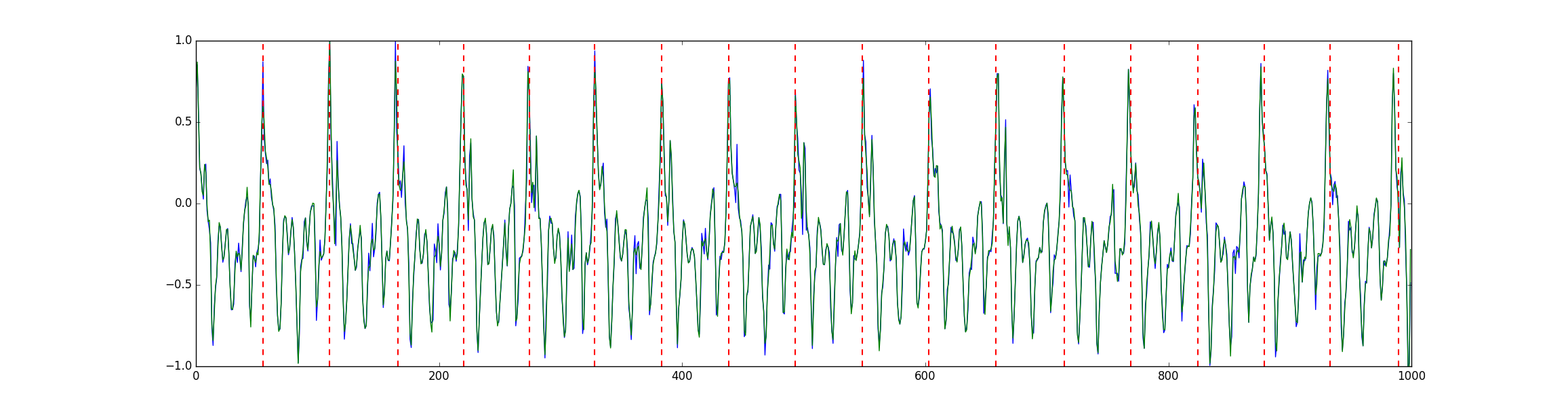}
\includegraphics[width=4in]{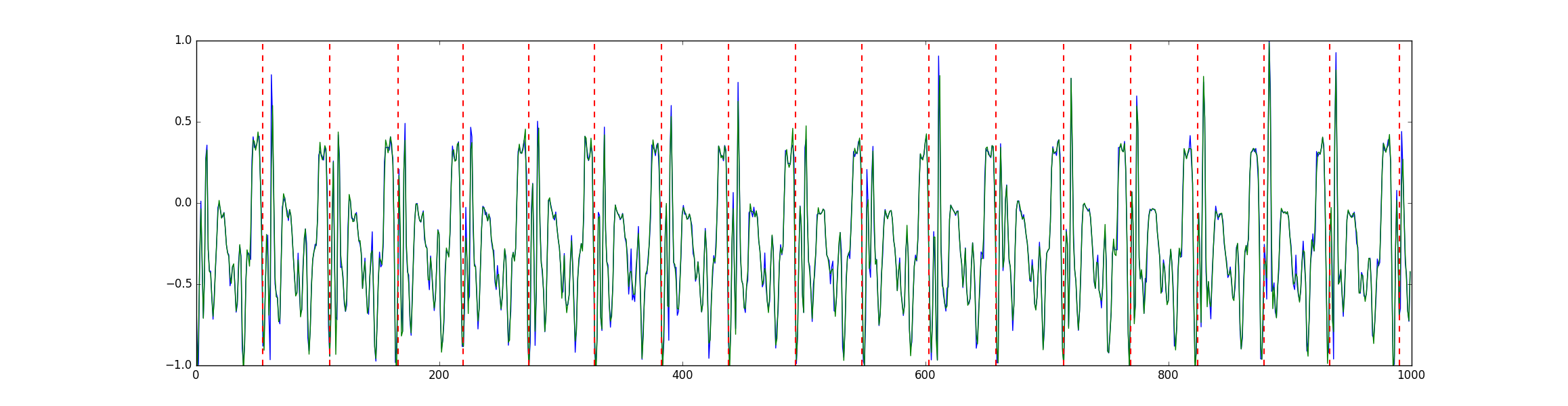}
\includegraphics[width=4in]{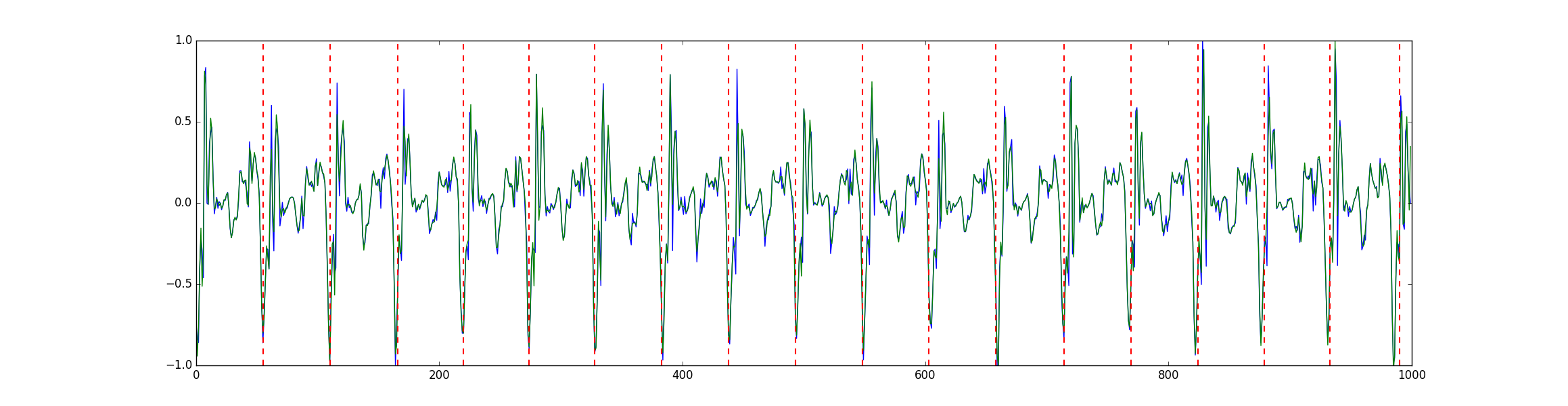}
\caption{Max tolerable deviation:0.15 Compression ratio: 0.05 RMSE: 0.02 (X-up, Y-mid, Z-bottom)}
\end{figure}

\begin{figure}[H]
\centering
\includegraphics[width=4in]{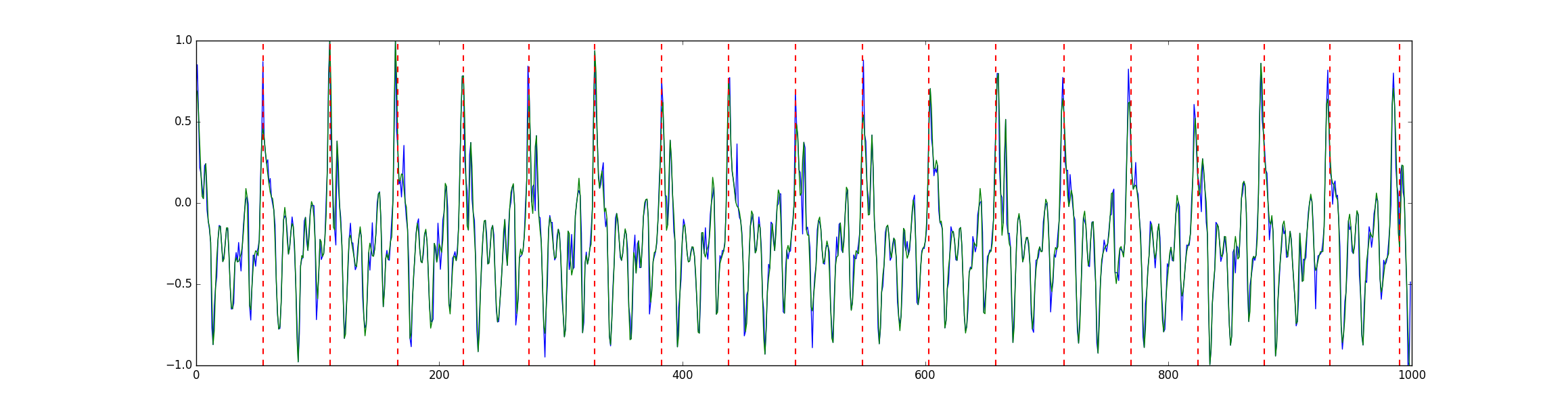}
\includegraphics[width=4in]{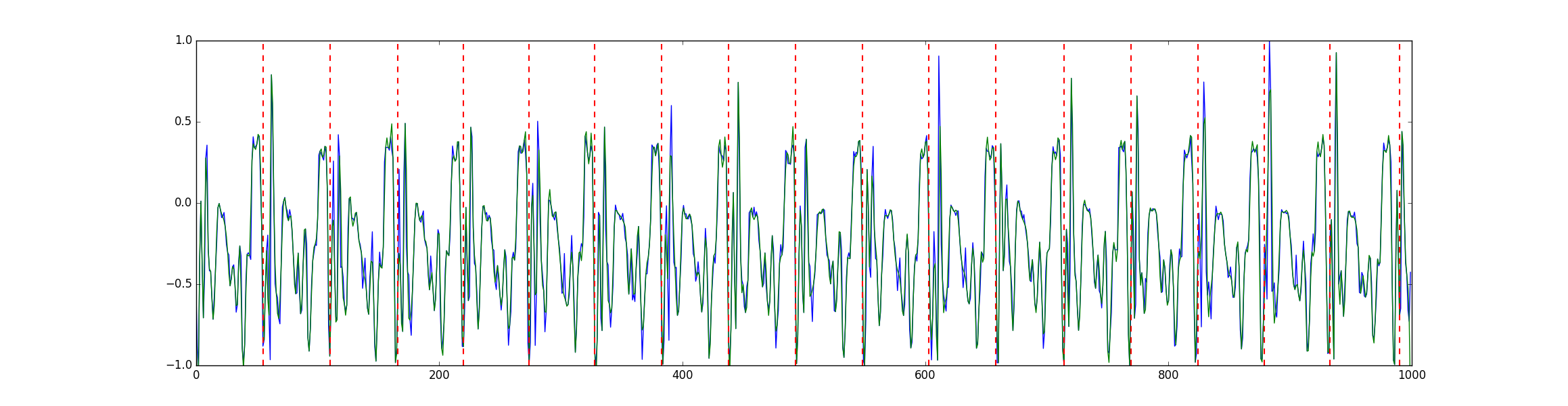}
\includegraphics[width=4in]{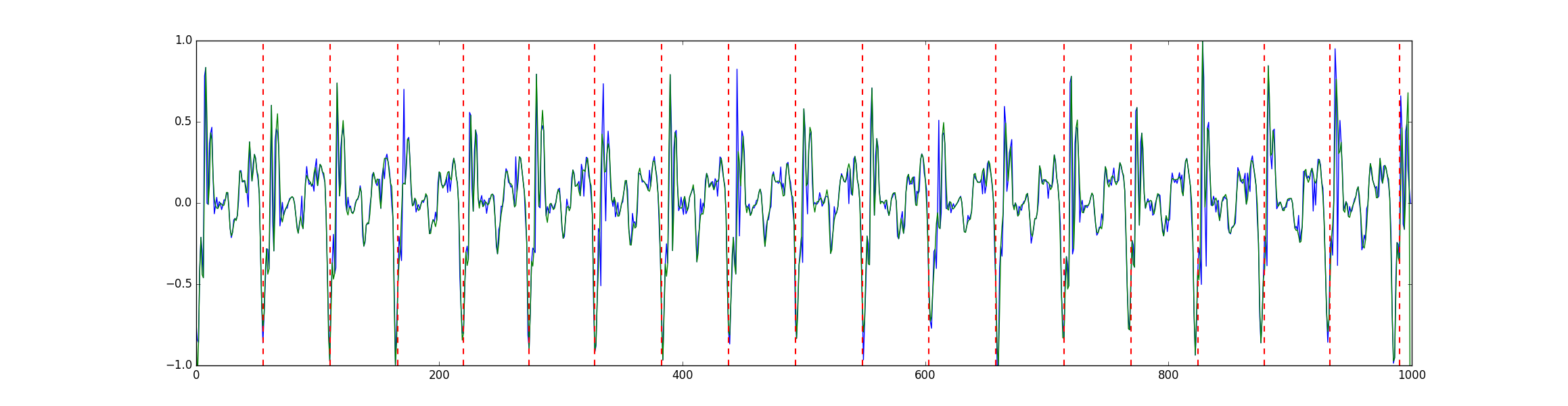}
\caption{Max tolerable deviation:0.2 Compression ratio: 0.03 RMSE: 0.05 (X-up, Y-mid, Z-bottom)}
\end{figure}

\section{Conclusion}
In this work, we propose a novel lossy compression algorithm for time series. The model is a combination of autoencoder and LSTM. And we apply an adaptive input window to adjust to local statistics of time series. Our work has two advantages:
\begin{itemize}
\item Both encoder and decoder can learn the long-term dependencies in time series, and the amount of information in the transmission can be reduced.
\item The adaptive input window size can make the algorithm ignore redundant information, and focus more on informative part of time series.
\end{itemize}
In the future, we plan to apply this algorithm to financial data.

\end{document}